\documentclass[]{ceurart}

\usepackage{booktabs, multirow, tabularx}
\newcolumntype{P}[1]{>{\centering\arraybackslash}p{#1}}
\newcolumntype{M}[1]{>{\centering\arraybackslash}m{#1}}

\begin{document}

\copyrightyear{2024}
\copyrightclause{Copyright for this paper by its authors.
  Use permitted under Creative Commons License Attribution 4.0
  International (CC BY 4.0).}

\conference{CLEF 2024: Conference and Labs of the Evaluation Forum, September 9-12, 2024, Grenoble, France}

\title{Multi-Label Plant Species Classification with Self-Supervised Vision Transformers}

\author[1]{Murilo Gustineli}[
orcid=0009-0003-9818-496X,
email=murilogustineli@gatech.edu,
url=https://murilogustineli.com,
]
\cormark[1]

\author[1]{Anthony Miyaguchi}[
orcid=0000-0002-9165-8718,
email=acmiyaguchi@gatech.edu,
]
\cormark[1]

\author[1]{Ian Stalter}[
email=istalter66@gmail.com,
]

\address[1]{Georgia Institute of Technology, North Ave NW, Atlanta, GA 30332}
\cortext[1]{Corresponding author.}

\begin{abstract}
    We present a transfer learning approach using a self-supervised Vision Transformer (DINOv2) for the PlantCLEF 2024 competition, focusing on the multi-label plant species classification.
    Our method leverages both base and fine-tuned DINOv2 models to extract generalized feature embeddings.
    We train classifiers to predict multiple plant species within a single image using these rich embeddings.
    To address the computational challenges of the large-scale dataset, we employ Spark for distributed data processing, ensuring efficient memory management and processing across a cluster of workers.
    Our data processing pipeline transforms images into grids of tiles, classifying each tile, and aggregating these predictions into a consolidated set of probabilities.
    Our results demonstrate the efficacy of combining transfer learning with advanced data processing techniques for multi-label image classification tasks.
    Our code is available at \href{https://github.com/dsgt-kaggle-clef/plantclef-2024}{github.com/dsgt-kaggle-clef/plantclef-2024}.
\end{abstract}


\maketitle

\section{Introduction}

The PlantCLEF 2024 challenge \cite{plantclef2024}, part of the LifeCLEF lab \cite{lifeclef2024} under the Conference and Labs of the Evaluation Forum (CLEF), aims to address the multi-label classification of plant species in high-resolution plot images.
This task presents unique challenges due to the shift between single-label training data (images of individual plants) and multi-label test data (images of vegetation plots) between the 2023 and 2024 editions.
The training dataset comprises approximately 7,800 species and 1.4 million images, totaling 281GiB, posing substantial computational challenges.

To address these challenges, we propose a transfer learning approach utilizing a Vision Transformer (ViT) model, specifically DINOv2 \cite{oquab2023dinov2} for feature extraction, and a linear classifier trained on the resulting embeddings.
We hypothesize that leveraging the rich ViT embedding space learned through pretrained self-supervised learning of massive datasets, combined with the appropriate post-processing steps, will be sufficient to achieve domain-expert performance on the multi-label classification task without the need to train models from scratch.

\begin{figure}[t!]
    \centering
    \includegraphics[width=\textwidth]{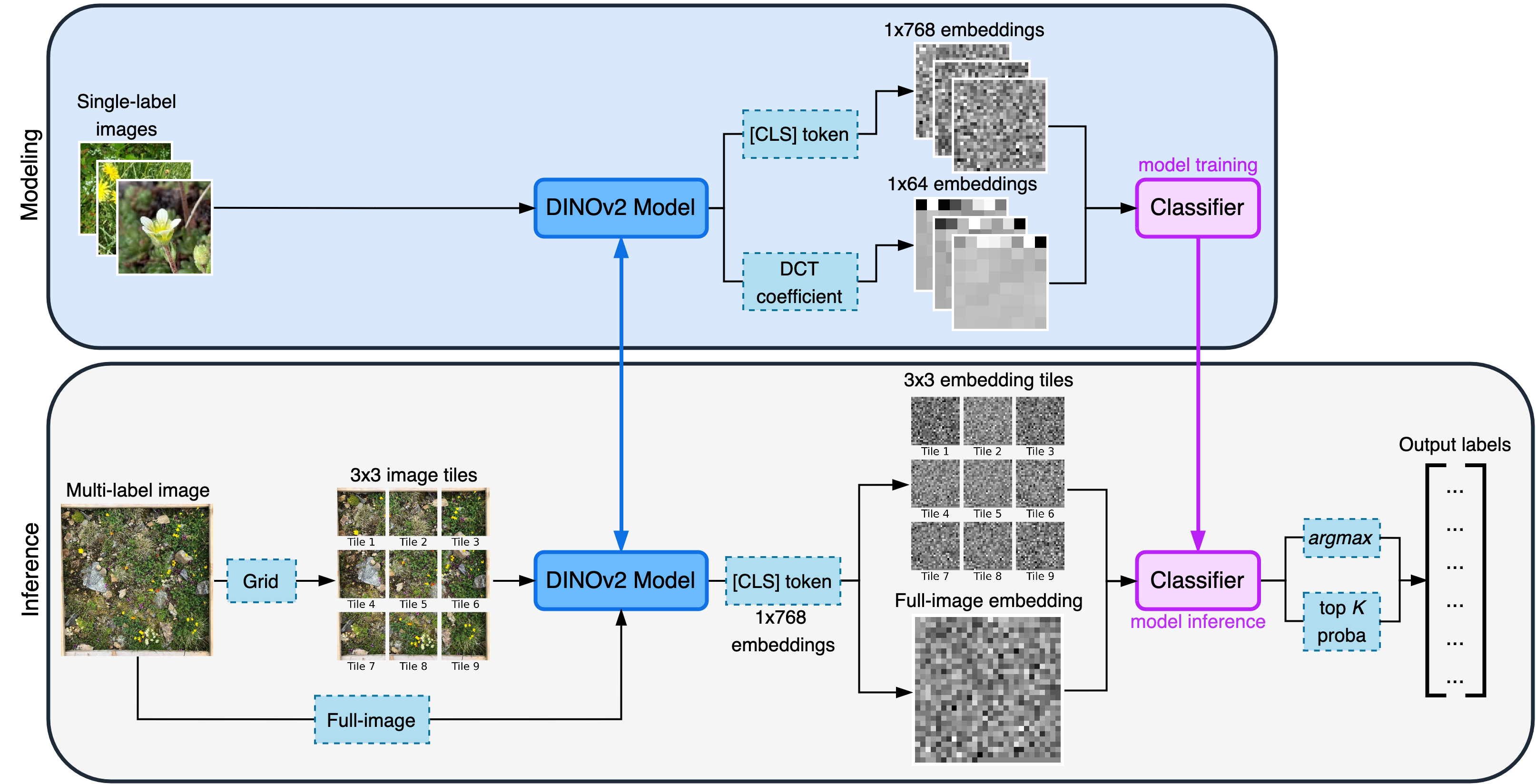}
    \caption{
    Overview of our proposed transfer learning method.
    In the modeling pipeline, we extract the DCT coefficient and [CLS] token embeddings from the single-label cropped and resized images using the base or fine-tuned DINOv2 model, and train a classifier on the embeddings.
    In the inference pipeline, DINOv2 extracts the [CLS] token embedding from each tile or full-image accordingly, followed by the trained classifier performing inference to obtain output species labels.
    }
\label{fig:model-diagram}
\end{figure}

\section{Overview}

Our approach leverages the embedding space learned by DINOv2 as a generalized feature representation of images, which are used to train models with higher bias (i.e., linear classifiers), as illustrated in Figure \ref{fig:model-diagram}.
DINOv2 learns robust feature representations by processing images as sequences of fixed-size patch tokens with an additional [CLS] token for classification tasks \cite{wu2023cls}.
These tokens serve as low-dimensional representations of the image patches, similar to words in a phrase for language models.
We train a linear classifier using the negative log-likelihood (NLL) loss on the single-label DCT coefficient and [CLS] token embeddings.
To address the multi-label classification problem, we perform inference using full-image and grid-based methods.


\subsection{DINOv2 Model Review}
\label{sec-dinov2-model}

DINOv2 is a state-of-the-art vision transformer encoder model, similar to BERT \cite{devlin2018bert}, pretrained without supervision on the LVD-142M dataset, a large collection of 142 million images.
Images are presented to the model as a sequence of fixed-size patches, which are linearly embedded.
A [CLS] token is added to the beginning of the sequence to facilitate classification tasks, and absolute position embeddings are included before feeding the sequence into the Transformer encoder layers.

The DINOv2 architecture comes in different sizes, each with its respective embedding dimensions: small (S) with 382, base (B) with 768, large (L) with 1024, and giant (g) with 1536.
We chose the ViT-B/14 (distilled) base model as it provides a balance of computational efficiency, performance, and feature representation suited for our use case, ensuring the model is powerful enough to extract meaningful features while remaining computationally feasible.
The base model produces embeddings with a fixed size of $\mathbb{R}^{257 \times 768}$, regardless of the input image dimensions.
This fixed size is due to the base model’s architecture: each image is divided into 256 fixed-size patches, and each patch is embedded into a 768-dimensional vector.
Additionally, a [CLS] token, also with a 768-dimensional embedding, is added to the sequence, resulting in a total of 257 vector embeddings per image \cite{dosovitskiy2020image}.
This base model does not include any fine-tuned heads and learns robust inner representations of images through pretraining, which can be utilized for downstream tasks.

The organizers provided two DINOv2 models based on the ViT-B/14 (distilled) architecture, each using a self-supervised learning method \cite{darcet2023vision}, trained with the \textbf{timm} library and hosted on Hugging Face.
The first model has its backbone frozen, fine-tuning only the classification head on new data, leveraging the robust feature extraction capabilities learned during pretraining.
The second model continues the first training but includes updates to both the backbone and the classification head, refining the feature representations throughout the network.
We utilize the second fine-tuned model to extract [CLS] token embeddings from the images, benefiting from its enhanced feature extraction capabilities.

\section{Methodology}

\begin{figure}[ht]
  \centering
  \includegraphics[width=\textwidth]{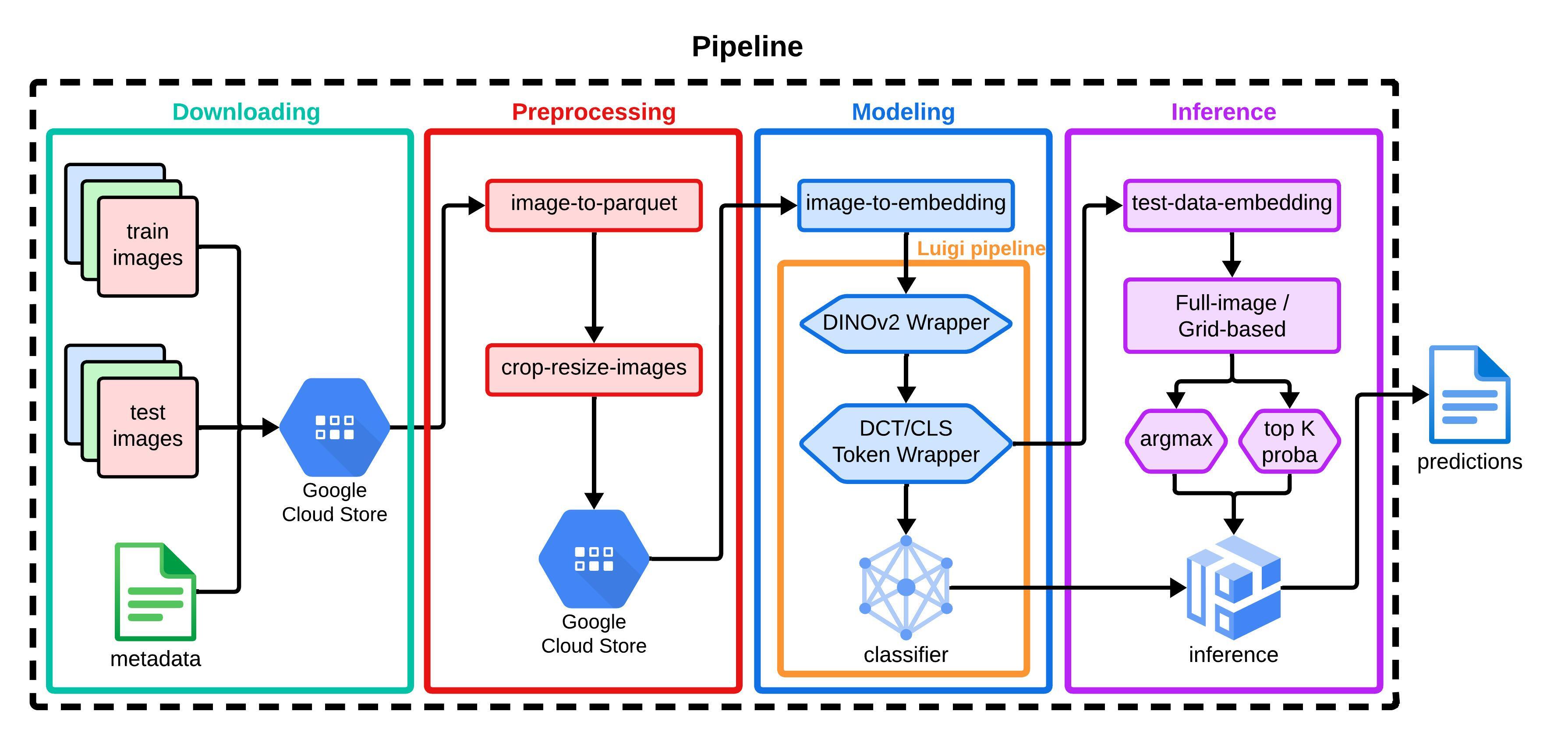}
  \caption{
  End-to-end pipeline of our proposed solution.
  The downloading module retrieves the training and test images, along with metadata, and stores them on Google Cloud Storage (GCS).
  The preprocessing module converts the images to binary data, crops and resizes them to $\mathcal{R}^{128\times 128}$ dimensions, and writes them as parquet files to GCS.
  The modeling module extracts embeddings using base and fine-tuned DINOv2 models and trains a linear classifier on the training embeddings. During inference, the trained classifier makes predictions on the test embeddings, formatting the results for leaderboard submission.
  }
  \label{fig:pipeline}
\end{figure}

We conducted experiments on the embedding datasets to maximize our performance on the public leaderboard.
Initially, we establish a baseline with minimal modifications, performing multi-class classification on the multi-label test dataset.
Subsequently, we introduced more complex inference approaches, such as grid-based image prediction for multi-label classification.

Our two main approaches were as follows: \textbf{(1) Extract embeddings} using both base and fine-tuned DINOv2 models from the cropped and resized single-label image dataset and the multi-label test dataset, train classifiers on the training embeddings, and perform classification on the test embeddings.
\textbf{(2) Perform inference} using the fine-tuned ViT model with both full-image and grid-based image prediction approaches.
We measure model performance using the metrics referenced in Section \ref{sec:class-imbalance}.
The results of our derived datasets and models are summarized in Table \ref{tab:linear-classifiers} and \ref{tab:inference}, respectively.

To address the computing and memory constraints of the large-scale training data, our solution leverages several technologies.
We use Google Cloud Platform (GCP) for computing and storage, Apache Spark for distributed data processing, Petastorm for distributed data loading, PyTorch Lightning for model training, and Weights and Biases for experiment tracking. 
Our primary compute resources include the \textbf{n2-standard-4} VM instance (4vCPU, 2 core, 16GiB memory) and the \textbf{g2-standard-8} GPU instance (8vCPU, 4 core, 32GiB memory), scaling up as needed for the dataset's magnitude.

Apache Spark was crucial to our entire pipeline, especially for preprocessing and modeling tasks.
PyTorch Lightning provides a high-level interface for our deep learning workflows, enabling efficient model training and hyperparameter tuning.
Our end-to-end pipeline comprises four main components:
downloading, preprocessing, modeling, and inference, illustrated in Figure \ref{fig:pipeline}.

\subsection{Downloading and Preprocessing}
We utilize aria2, a lightweight multi-protocol and multi-source command-line download utility that facilitates fast and reliable downloading of large datasets.
The images and metadata files were downloaded and stored in Google Cloud Storage for subsequent preprocessing.

The preprocessing phase involves two main steps: converting images to Apache Parquet format and performing cropping and resizing operations.
We concatenate image data in batches to optimize cloud computation, as reading millions of images incurs significant network overhead.
Columnar data formats like Parquet efficiently represent data for batch processing, handling both binary and metadata.

\begin{figure}[ht]
    \centering
    \includegraphics[width=0.8\textwidth]{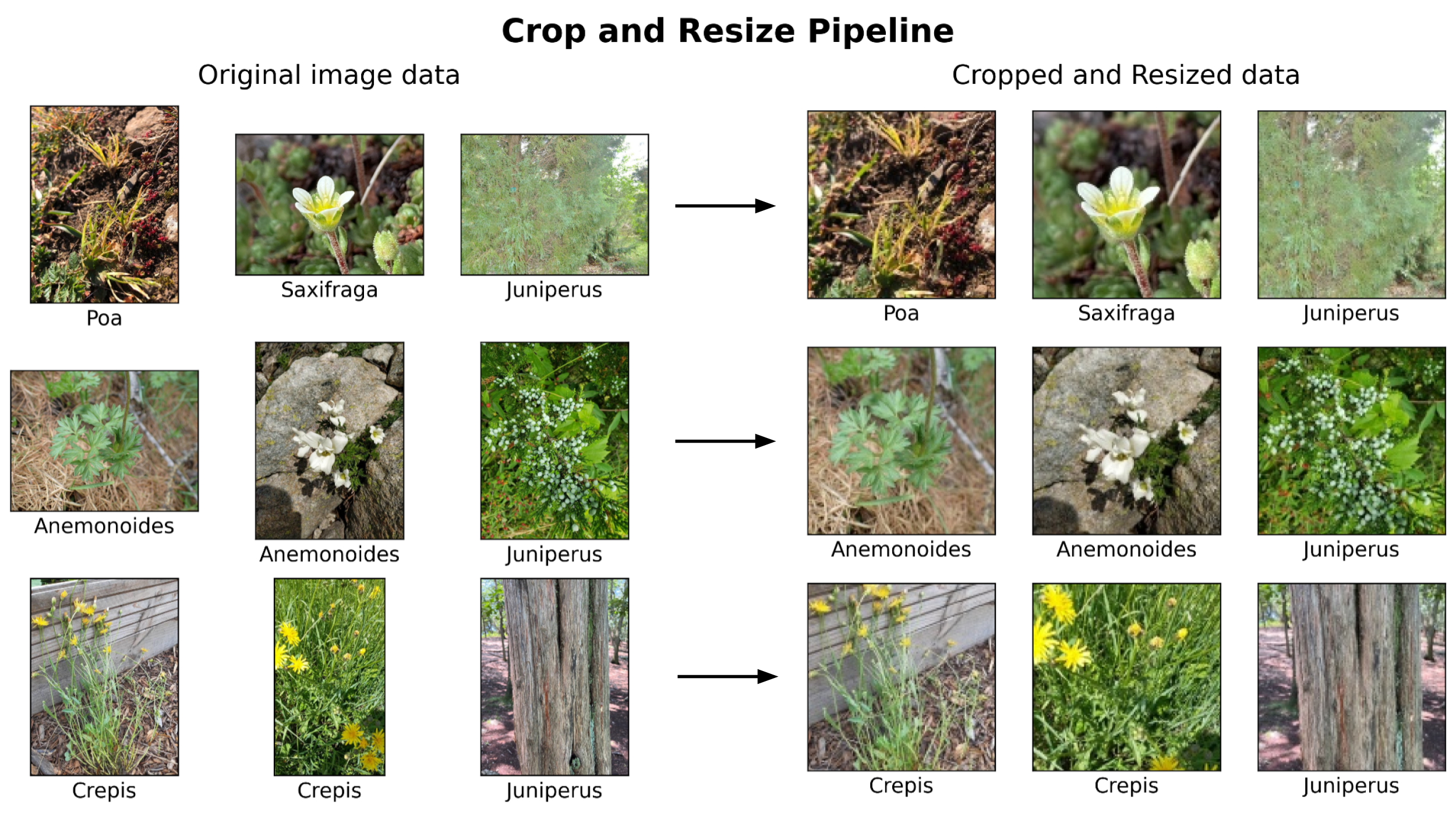}
    \caption{
    Comparison of original images with $\mathcal{R}^{128 \times 128}$ cropped and resized squared images. The original images have a minimum resolution of 800 pixels on the longest side, allowing for the use of high-resolution classification models and potentially improving the prediction of small plants in large vegetative plots.
    }
    \label{fig:crop-resize}
\end{figure}

Many images in the dataset are rectangular and inconsistent in resolution.
To improve processing efficiency, we crop and resize all images to ensure the subject remains in focus, as shown in Figure \ref{fig:crop-resize}.
Each image is cropped to a square centered at the midpoint and resized to $\mathcal{R}^{128\times 128}$ pixels.
This step reduced the dataset size from 281GiB to approximately 15GiB, achieving more than an order of magnitude reduction. 
This reduction facilitated faster embedding extraction using DINOv2 and decreased computational load during training and inference.
We choose a relatively small, square dimension to allow for constructing a multi-class dataset from collages of smaller images.

We preprocess each image into a grid of tiles and extract DINOv2 features for each tile (Figure \ref{fig:model-diagram}).
Using the base DINOv2 model, we extract full-size embeddings $\mathcal{R}^{257\times 768}$ from each image, including DCT coefficients of the tile tokens and the [CLS] token.
With the fine-tuned DINOv2 model, we focus on extracting the [CLS] token embeddings.
We create the following embedding datasets in Table \ref{tab:dino-models}.

\textbf{Base DINO DCT.}
The base DINOv2 model extracts embeddings from the cropped and resized single-label image dataset.
The embeddings have dimensions of $\mathcal{R}^{257\times 768}$.
We apply the DCT algorithm for dimensionality reduction with an 8x8 filter size, resulting in a $\mathcal{R}^{1\times 64}$ tensor.
The DCT captures multi-dimensional low-rank structures in frequency and is known for its compressive properties on datasets, such as JPEG and MP3 for images and audio respectively \cite{ahmed1974discrete}.
It is a data-independent transformation that runs in $O(n\log n)$ time, with accessible and efficient implementations, unlike data-dependent transformations like singular value decomposition (SVD) which requires $O(n^3)$ eigen-decompositions.
The DCT can identify periodicity within the 2D patch tokens and a low-frequency space useful for downstream tasks with minimal overhead.

\textbf{Base DINO [CLS] token.}
The base DINOv2 model extracts the [CLS] token embeddings from the cropped and resized dataset.
The [CLS] token is a special token added to the input sequence of the Vision Transformer, aggregating information from all tiles and providing a generalized feature representation of the entire image for classification tasks.
The resulting [CLS] token embedding is a tensor of $\mathcal{R}^{1\times 768}$ dimensions.

\textbf{Fine-tuned DINO.}
We use the fine-tuned model \textbf{dinov2-onlyclassifier-then-all} discussed in Section \ref{sec-dinov2-model} to extract [CLS] token embeddings from the cropped and resized single-label image dataset, resulting in a tensor of shape $\mathcal{R}^{1\times 768}$.

\begin{table}[!t]
\centering
\caption{
An overview of the embedding datasets created by utilizing the DINOv2  model to extract embeddings from the cropped and resized single-label image dataset.
\textbf{facebook-dinov2-base} refers to the base off-the-shelf DINOv2 model, and \textbf{dinov2-onlyclassifier-then-all} refers to the fine-tuned ViT model provided by the organizers.
}
\label{tab:dino-models}
\begin{tabular}{*5l} \toprule
\textbf{DINOv2 Model} & \textbf{Dataset created} & \textbf{Size} \\ \midrule
facebook-dinov2-base & Base DINO DCT & 403.85MiB \\
facebook-dinov2-base & Base DINO [CLS] token & 4.09GiB \\
dinov2-onlyclassifier-then-all (non-frozen) & Fine-tuned DINO & 4.09GiB \\ \bottomrule
\end{tabular}
\end{table}

\subsection{Modeling and Inference}

We train linear classifiers on both DCT-reduced embeddings and [CLS] token embeddings using PyTorch Lightning and the negative log-likelihood loss.
For inference, we performed multi-class classification on full-size images to predict a single plant species per image.

To address multi-label classification, each test image is divided into a grid of tiles, with embeddings extracted for each tile using the fine-tuned DINOv2 model. We then perform inference on these tile embeddings using \textit{argmax} and top $K$ probabilities for prediction aggregation.

Our inference workflow employs a Luigi \cite{Luigi2023} task, which processes the image prediction output by extracting species IDs corresponding to the top $L$ probabilities for each image.
Duplicates are removed, and order is preserved by converting to a set and sorting by their original appearance in the logits list.
The unique species IDs are compiled into a structured dataframe, with each record corresponding to an image, formatted for submission, and written to a CSV file.

\begin{figure}[h!]
    \centering
    \includegraphics[width=0.80\textwidth]{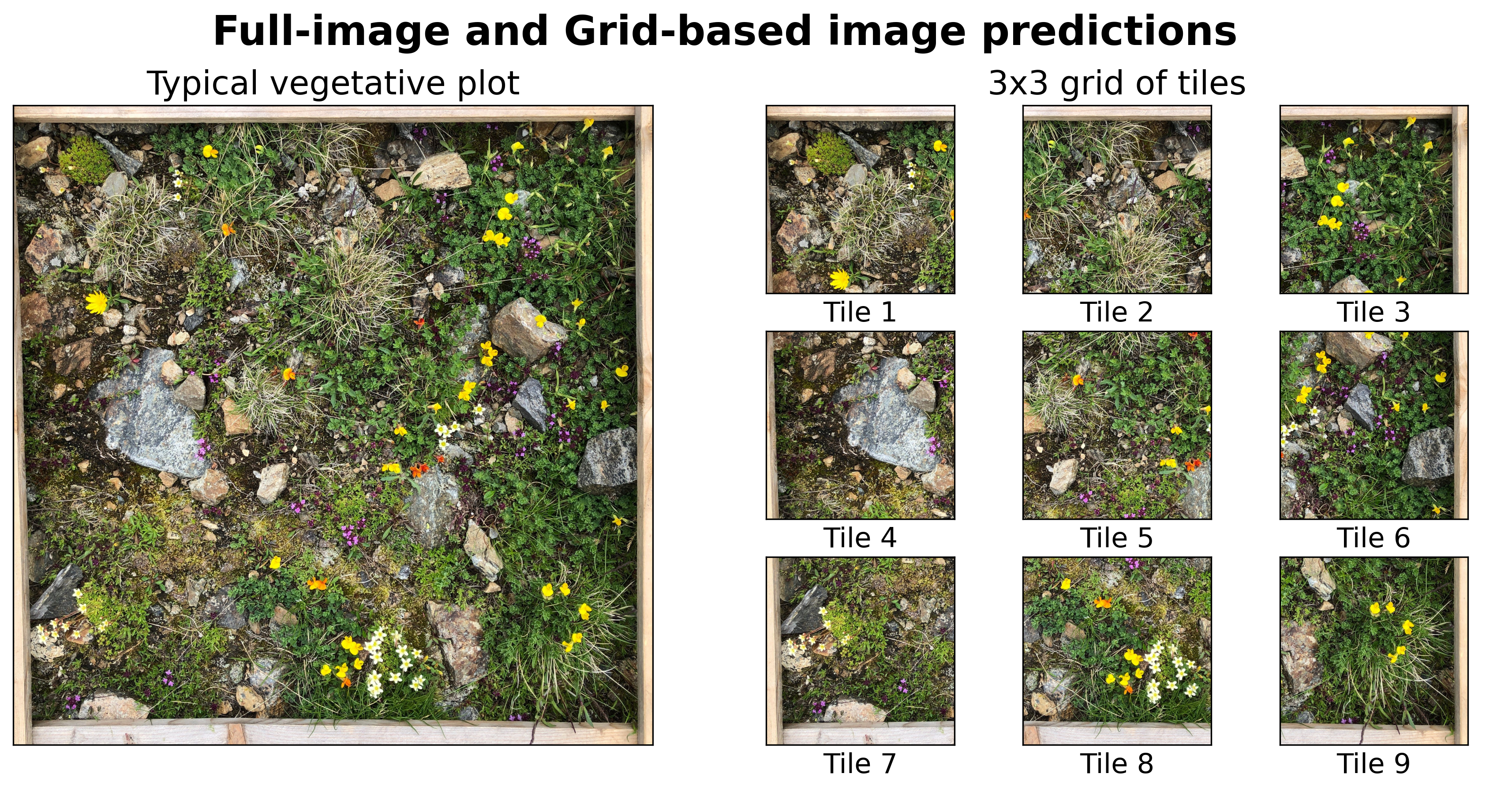}
    \caption{
    Comparison of full-image prediction and grid-based image prediction.
    The left plot shows a typical vegetative plot from the test set, where a botanist recorded 8 species: \textit{Cardamine resedifolia L., Festuca airoides Lam., Pilosella breviscapa (DC.) Soják, Lotus alpinus (Ser.) Schleich. ex Ramond, Poa alpina L., Saxifraga moschata Wulfen, Scorzoneroides pyrenaica (Gouan) Holub, and Thymus nervosus J.Gay ex Willk}.
    The right plot illustrates the same image divided into a $3\times 3$ grid, demonstrating the grid-based approach for species classification by processing each tile independently.
    }
\label{fig:original-grid}
\end{figure}

We employ two distinct approaches to image prediction on the test dataset, as shown in Figure \ref{fig:original-grid}: \textbf{full-image} and \textbf{grid-based image prediction}.
The test dataset is not cropped and resized to preserve the high quality of multi-label images.

\textbf{Full-Image Prediction.} In this approach, the entire test image is processed in its original dimension.
The fine-tuned ViT model evaluates the image and outputs probabilities for each of the 7806 plant species classes.
We then extract the top 20 probabilities, representing the most likely species present, and map these probabilities to their corresponding species IDs.

\textbf{Grid-based Image Prediction.} We segment the image into an $N\times N$ grid of tiles, resulting in $M$ tiles.
Each tile is independently processed using the fine-tuned ViT model, which outputs probabilities for each species class.
We then select the top $K$ probabilities (default $K=10$) for each tile and map them to their respective species classes.
This produces a nested array of prediction lists per tile, which we flatten into a single array, further limiting to the top $L$ probabilities (default $L=5$) from each tile.
For example, a $3\times 3$ grid yields nine tiles, each with ten top probabilities, selecting the top five logits in each tile, totaling 45 species IDs and probability mappings.

Let $P_{i,j}$ be the probability of species $i$ in tile $j$. For the \textbf{argmax} approach, we select:
\begin{equation}
\hat{y}_{j} =\underset{i}{\text{argmax}} P_{i,j}
\end{equation}
For the \textbf{top $K$ probability} approach, we aggregate the top $K$ logits per tile, and  select the top $L$ species across all tiles, where $L=5$, and $M$ is the total number of tiles:
\begin{equation}
\hat{Y} =\text{Top-} L\left(\bigcup _{j=1}^{M}\text{Top-} K( P_{:,j})\right)
\end{equation}

We generate a consolidated array of predictions whether processing the entire image or employing the grid-based approach. This array, a collection of species ID and probability mappings, is sorted in descending order based on the probability scores. This prioritization helps accurately identify the most likely species in each image tile.

\subsection{Class Imbalance}
\label{sec:class-imbalance}
The training dataset exhibits significant class imbalance, with a highly right-skewed distribution of species.
Of the approximately 7,800 species, nearly half have fewer than 100 images.
To mitigate this imbalance, we select a subset of plant species with at least 100 images for training
While this approach may overlook rare species, we hypothesize that focusing on more represented species will increase the confidence in species identification.

By selecting species with a minimum of 100 images, we aim to improve the balance between precision and recall for the included species, directly impacting the Macro F1 Per Species and Macro F1 Per Plot metrics.
While this may lead to a lower recall for rare species in the test set, it enhances the precision for the more common species, resulting in higher Macro F1 scores for those species.
For the Micro F1 score, which is more sensitive to the performance of common species due to the overall count of true positives, false positives, and false negatives, our approach is likely to result in higher scores as well.

\subsection{Evaluation Metrics}

The metrics used to evaluate our model are Macro F1 Averaged Per Plot, Macro F1 Averaged Per Species, and Micro F1 scores.

The F1 score is the harmonic mean of precision and recall, defined as:
\begin{equation}
    F1=2\cdot \frac{P\cdot R}{P+R}
\end{equation}
where $P$ and $R$ denote precision and recall, respectively.

The Macro F1 Averaged Per Plot and Macro F1 Averaged Per Species are calculated as follows:
\begin{equation}
    \text{Macro F1 Per Plot} =\frac{1}{N}\sum _{i=1}^{N} F1( y_{i} ,\hat{y}_{i})
\end{equation}
\begin{equation}
    \text{Macro F1 Per Species} =\frac{1}{C}\sum _{c=1}^{C} F1( y_{c} ,\hat{y}_{c})
\end{equation}
where $N$ is the number of plots, $C$ is the number of species, $y_i$ is the true label for plot $i$, and $\hat{y}_i$ is the predicted label for plot $i$.

The Micro F1 score aggregates the contributions of all classes to compute the average F1 score:
\begin{equation}
    \text{Micro F1} =\frac{2\sum _{c=1}^{C} TP_{c}}{2\sum _{c=1}^{C} TP_{c} +\sum _{c=1}^{C} FP_{c} +\sum _{c=1}^{C} FN_{c}}
\end{equation}

\section{Results}

We present our best results on the public and private leaderboards.
Our best model utilizes a linear classifier trained on the fine-tuned DINO embeddings (Table \ref{tab:dino-models}) and performs inference using a grid-based approach with argmax logit per tile.
We achieved public scores of 20.77 for Macro F1 Averaged Per Plot, 47.42 for Macro F1 Averaged Per Species, and 19.67 for Micro F1, as shown in Tables \ref{tab:team-results} and \ref{tab:linear-classifiers}.

\begin{table}[h]
\centering
\caption{
A summary of the top 3 best scores in the public and private leaderboards. Our solution achieved the third-highest score in both leaderboards.
}
\label{tab:team-results}
\begin{tabular}{lllM{0.6in}M{0.7in}M{0.5in}} \toprule
\textbf{Leaderboard} & \textbf{Team name} & \textbf{Rank} & \textbf{MacroF1 Averaged Per Plot} & \textbf{MacroF1 Averaged Per Species} & \textbf{MicroF1} \\ \midrule
\multirow{3}{*}{Public} & Atlantic & 1 & 29.62 & 51.2 & 30.01 \\ 
    & NEUON AI & 2 & 23.01 & 46.2 & 20.84 \\ 
    & DS@GT-LifeCLEF (Ours) & 3 & 20.77 & 47.42 & 19.67 \\ \cmidrule(r){1-6}
\multirow{3}{*}{Private} & Atlantic & 1 & 28.73 & 45.76 & 29.57 \\ 
    & NEUON AI & 2 & 21.31 & 34.69 & 20.75 \\ 
    & DS@GT-LifeCLEF (Ours) & 3 & 19.04 & 32.4 & 19 \\ \bottomrule
\end{tabular}
\end{table}


The linear classifier trained on the fine-tuned DINOv2 embeddings consistently outperforms base DINOv2 models across all metrics, demonstrating enhanced feature representation from additional training (Table \ref{tab:linear-classifiers}).
The base DINOv2 models show relatively low performance, with the model trained on the [CLS] token embeddings performing slightly better than the DCT embeddings.
The lower performance of the DCT embeddings is due to their reduced dimensionality $\mathcal{R}^{1\times 64}$ compared to the [CLS] token embeddings $\mathcal{R}^{1\times 768}$, leading to information loss.
While base [CLS] token embeddings are more effective than DCT embeddings, both are less effective than fine-tuned model embeddings.

\begin{table}[h]
\centering
\caption{
Summary of the linear classifiers trained on the embedding datasets. All classifiers are single-layer linear networks implemented with PyTorch's \texttt{nn.Linear()} and trained using the negative log-likelihood loss.
The embedding dataset used to train the classifiers and the inference method significantly impact the results, with more capable models extracting richer features that improve classification performance.
}
\label{tab:linear-classifiers}
\begin{tabular}{lM{1.3in}M{0.6in}M{0.7in}M{0.5in}}
\toprule
\textbf{Dataset} & \textbf{Inference method} & \textbf{MacroF1 Averaged Per Plot} & \textbf{MacroF1 Averaged Per Species} & \textbf{MicroF1} \\ \midrule
base dino dct & \multirow{2}{*}{Multi-class, full-image}  & 0.28 & 33.33 & 0.1 \\ \cmidrule(r){1-1} \cmidrule(r){3-5}
base dino cls token &                                     & 2.06 & 31.46 & 1.62 \\ \cmidrule(r){1-5}
\multirow{3}{*}[-3ex]{fine-tuned dino} & Multi-class, full-image & 8.47 & 34.39 & 7.04 \\ \cmidrule(r){2-5}
    & Multi-class, grid-based, argmax logit per tile & \textbf{20.77} & \textbf{47.42} & \textbf{19.67} \\ \cmidrule(r){2-5}
    & Multi-label, grid-based, top 5 species per tile & 20.62 & 42.31 & 19.07 \\
\bottomrule
\end{tabular}
\end{table}

The multi-class, full-image inference method with the linear classifier trained on fine-tuned DINOv2 embeddings shows significant improvement over the base models.
For a more sophisticated approach, we utilize a grid-based inference method to improve multi-label classification capability, employing two strategies: \textbf{argmax logit per tile} and \textbf{top 5 species per tile}.
The grid-based approach with argmax logit per tile achieves the highest scores across all metrics.
The grid-based approach with top 5 species per tile also performs well with slightly lower performance due to less confident predictions introducing noise and ambiguity, and the complexity of aggregating multiple logits, resulting in suboptimal predictions.

For inference using the fine-tuned model, we found that a $3\times 3$ grid size struck a balance between computational efficiency and species distribution, having experimented with $2\times 2$ and $5\times 5$ grid sizes without observing significant improvements, as shown in Table \ref{tab:inference}.
We chose 5 species per tile to maximize the Macro F1 Score Per Species, focusing on the most represented species in the dataset.
However, this approach leaves room for improvement, particularly for rare species that are underrepresented.

\begin{table}[h]
\centering
\caption{
An overview of the inference experiments using the fine-tuned ViT model on the test embedding dataset.
}
\label{tab:inference}
\begin{tabular}{llM{0.6in}M{0.7in}M{0.5in}} \toprule
\textbf{Inference method} & \textbf{Prediction method} & \textbf{MacroF1 Averaged Per Plot} & \textbf{MacroF1 Averaged Per Species} & \textbf{MicroF1} \\ \midrule
\multirow{2}{*}[-0.5ex]{Multi-label, full-image} & Top 5 species & 9.88 & 43.87 & 8.96 \\ \cmidrule(r){2-5}
    & Top 20 species          & 7.88  & 33.64 & 7.98  \\ \cmidrule(r){1-5}
\multirow{4}{*}[-1.5ex]{Multi-label, grid-based} & Top 5 species, 3x3 grid & \textbf{17.76} & \textbf{44.76} & \textbf{14.39} \\ \cmidrule(r){2-5}
    & Top 3 species, 3x3 grid & 16.30 & 43.98 & 12.26 \\ \cmidrule(r){2-5}
    & Top 5 species, 5x5 grid & 17.71 & 41.84 & 14.36 \\ \cmidrule(r){2-5}
    & Top 5 species, 2x2 grid & 15.41 & 42.64 & 12.92 \\
\bottomrule
\end{tabular}
\end{table}

\section{Discussion}

By leveraging transfer learning, we use embeddings from both base and fine-tuned DINOv2 models to train linear classifiers, addressing the challenge of single-label training data versus multi-label test data.
Our preprocessing pipeline manages the large-scale dataset by converting images to Parquet format, reducing dimensionality through cropping and resizing, and decreasing computational load for extracting embeddings and making inferences.

We demonstrate the effectiveness of using DCT for dimensionality reduction and the [CLS] token for generalized feature representation.
The fine-tuned DINOv2 model provides enhanced feature representations crucial for multi-label classification.
Our grid-based inference approach with argmax and top $K$ probability aggregation enables accurate prediction of multiple species within each image.

We did not test the base DINOv2 in a grid-based manner or the fine-tuned DINOv2 with full-image inference due to computational constraints and our hypothesis that fine-tuning would yield better results when exploiting local features through tiling.
While capturing the top 5 logits per tile provides broader species identification, it also introduces less confident predictions, complicating final aggregation.
Conversely, the argmax method directly selects the highest probability, leading to better predictions.

\subsection{Species Representation in Embedding Spaces}

\begin{figure}[ht]
    \centering
    \includegraphics[width=1\textwidth]{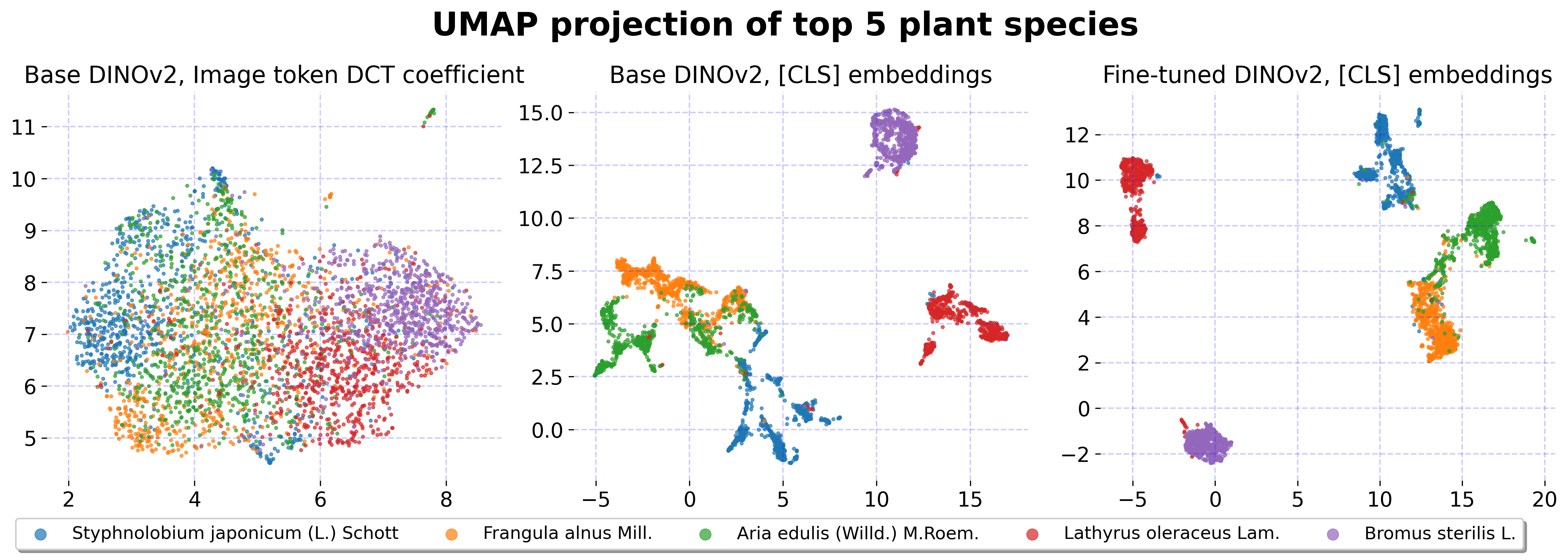}
    \caption{
    UMAP projections of the top 5 plant species with the highest number of images.
    The fine-tuned model's embeddings exhibit better spatial separation, highlighting their effectiveness as feature representations.
    }
\label{fig:umap-cluster}
\end{figure}

In Figure \ref{fig:umap-cluster}, we visualize the spatial separation between species using the embeddings extracted from the base and fine-tuned ViTs.
The UMAP \cite{mcinnes2018umap} projections of the top five plant species illustrate the importance of training a model to create a discriminating feature space and the challenge of obtaining unique representations without a learning mechanism like the DCT coefficient.

The image token DCT coefficients exhibit the poorest species separation among the three approaches.
Although effective for dimensionality reduction and capturing low-frequency information, DCT does not inherently learn discriminative features for classification tasks.
In contrast, the base DINOv2 [CLS] token embeddings show better species separation, as the [CLS] token aggregates information from all image patches, providing a more comprehensive representation of the entire image.

The fine-tuned DINOv2 model's [CLS] embeddings demonstrate the clearest species separation.
The additional training on the specific task data refines the model's feature representations, making them more relevant and discriminative for classification.
Fine-tuning effectively bridges the gap between general-purpose feature extraction and task-specific discriminative power, resulting in more accurate and reliable species classification, as shown in Table \ref{tab:linear-classifiers}.

\section{Future Work}

One direction for future work is to generate a collage dataset by tiling individual species that are likely to co-occur.
We propose a collaborative filtering approach to address the multi-label problem, leveraging Locality Sensitive Hashing (LSH) \cite{dasgupta2011fastLSH}, Approximate Nearest Neighbor Search (ANN) \cite{li2019approximate}, and Alternating Least Squares (ALS) \cite{takacs2012alternating}.
LSH reduces dimensions and detects similarity by hashing similar items into the same “buckets” with high probability.
ANN then efficiently finds the nearest neighbors within this subset, and ALS generates recommendations by combining scores with geographic proximity from LSH and ANN, ranking and recommending images based on their geographic closeness and similarity.
This method could generate images for direct use in multi-label learning.

\section{Conclusion}

We present a robust multi-label plant species classification approach using self-supervised Vision Transformer (DINOv2) models.
Our study highlights the potential of combining self-supervised learning and transfer learning with data processing techniques to tackle large-scale biodiversity challenges.
We offer a scalable solution for multi-label image classification using only single-label training data.
Future work could enhance model training and inference by integrating additional data augmentation techniques, experimenting with various grid sizes, exploring other dimensionality reduction methods, and utilizing alternative loss functions such as binary cross-entropy and asymmetric loss \cite{ridnik2021asymmetric}.
Additionally, developing more sophisticated aggregation strategies for multi-label prediction could further improve classification performance.
Our code is available at \href{https://github.com/dsgt-kaggle-clef/plantclef-2024}{github.com/dsgt-kaggle-clef/plantclef-2024}.

\begin{acknowledgments}

We want to thank the Data Science at Georgia Tech (DS@GT)-CLEF group for cloud infrastructure and their support, and the organizers of PlantCLEF and LifeCLEF for hosting the competition.
\end{acknowledgments}

\bibliography{main}

\end{document}